\documentclass[manuscript]{acmart}

\AtBeginDocument{%
  \providecommand\BibTeX{{%
    \normalfont B\kern-0.5em{\scshape i\kern-0.25em b}\kern-0.8em\TeX}}}

\copyrightyear{2023}
\acmYear{2023}
\setcopyright{acmlicensed}\acmConference[RecSys '23]{Seventeenth ACM Conference on Recommender Systems}{September 18--22, 2023}{Singapore, Singapore}
\acmBooktitle{Seventeenth ACM Conference on Recommender Systems (RecSys '23), September 18--22, 2023, Singapore, Singapore}
\acmPrice{15.00}
\acmDOI{10.1145/3604915.3608816}
\acmISBN{979-8-4007-0241-9/23/09}
\begin{document}

\title{Using Learnable Physics for Real-Time Exercise Form Recommendations}


\author{Abhishek Jaiswal}
\email{abhi.jaiswal44@gmail.com}
\orcid{0000-0003-4795-6205}

\affiliation{%
  \institution{Indian Institute of Technology, Kanpur}
  \streetaddress{Kalyanpur}
  \city{Kanpur}
  \state{Uttar Pradesh}
  \country{India}
  \postcode{208016}
}

\author{Gautam Chauhan}
\email{gautamchauhan0412@gmail.com}
\orcid{0000-0003-4957-9140}

\affiliation{%
  \institution{Indian Institute of Technology, Kanpur}
  \streetaddress{Kalyanpur}
  \city{Kanpur}
  \state{Uttar Pradesh}
  \country{India}
  \postcode{208016}
}

\author{Nisheeth Srivastava}
\email{nisheeths@gmail.com}
\orcid{0000-0001-9272-8418}

\affiliation{%
  \institution{Indian Institute of Technology, Kanpur}
  \streetaddress{Kalyanpur}
  \city{Kanpur}
  \state{Uttar Pradesh}
  \country{India}
  \postcode{208016}
}


\begin{abstract}
  
Good posture and form are essential for safe and productive exercising. Even in gym settings, trainers may not be readily available for feedback. Rehabilitation therapies and fitness workouts can thus benefit from recommender systems that provide real-time evaluation. In this paper, we present an algorithmic pipeline that can diagnose problems in exercises technique and offer corrective recommendations, with high sensitivity and specificity, in real-time. We use MediaPipe for pose recognition, count repetitions using peak-prominence detection, and use a learnable physics simulator to track motion evolution for each exercise. A test video is diagnosed based on deviations from the prototypical learned motion using statistical learning. The system is evaluated on six full and upper body exercises.These real-time recommendations, counseled via low-cost equipment like smartphones, will allow exercisers to rectify potential mistakes making self-practice feasible while reducing the risk of workout injuries.

\end{abstract}



\maketitle

\section{Introduction}
Sedentary lifestyles and physical inactivity are prominent risk factors for cardiovascular diseases worldwide. Evidence also suggests that physical activity has dipped considerably over time~\cite{ozemek2019global}. Exercise improves life expectancy and has an effective therapeutic impact on physical and mental health \cite{jimenez2020physical}. Assistance in performing physical exercises~\cite{fletcher2018promoting}
, therefore, has a vital role to play in improving health on a global scale.

Expert supervision in performing exercises is not readily available and even if present, self practice needs further assistance. Consequently, digital technologies have become instrumental in improving accessibility to expert supervision. In gym settings, 
Sensor-based methods~\cite{ velloso2013qualitative, spina2013copdtrainer} have been prolifically used for pose assessment focusing on exercise detection \cite{chang2007tracking, vseketa2015real, seeger2011myhealthassistant}, rep counting \cite{soro2019recognition, spina2013copdtrainer}, incorrect pose diagnosis \cite{yurtman2014automated, giggins2014rehabilitation, lee2020automatic, kowsar2016detecting} and suggestions \cite{zhao2014realtime,velloso2013qualitative,spina2013copdtrainer}. However, sensors can be obtrusive, expensive, and difficult to calibrate correctly, and so may be best suited for high-performance settings~\cite{khurana2018gymcam}. More appropriate for less intensive settings, vision-based methods \cite{wang2019ai, liu2020posture, gharasuie2021performance, wang2021monocular} have recently gained prominence due to improvements in deep learning techniques and mobile camera technology. This direction of research is very promising because it allows for the possibility of entirely sensor-free tracking of exercise performance. 

At present, however, such proposals face considerable difficulties. Most vision-based approaches to exercise tracking work with predetermined heuristic parameters which vary across exercises and participants, requiring considerable hand-crafting~\cite{liu2020posture, chen2020pose}. While vision-based approaches to exercise type recognition and rep-counting are plentiful, approaches that seek to track exercise form are limited to simple upper body exercises with relatively little body movement~\cite{liu2020posture, kowsar2016detecting, chen2020pose}. Further, most such approaches offer exercise diagnoses retrospectively after processing entire recorded exercise sessions~\cite{khurana2018gymcam, soro2019recognition}.

Taking the conventional idea of recommendation from collaborative filtering a step further, we frame the task of exercise supervision as that of recommending correct forms. To this end, we identify the over-general nature of the deep learning architectures used in vision-based exercise tracking pipelines as a key problem blocking progress in this area. Rather than use generic neural network architectures, we propose using a specific variety of neural networks, specifically designed to learn relationships between physical objects, as the base inference engine in such recommender systems. Using one such architecture - Interaction Networks~\cite{battaglia2016interaction} - we describe a novel recommender system for real-time exercise form correction in this paper. We show that our solution works with very high sensitivity and specificity for a wide variety of full-body and upper-body exercises. 

Section \ref{section:RelatedWork} places our proposal in line with existing approaches of exercise-specific recommendations.
We explain the working of our physics inference model in Section \ref{section:INgeneral} and the components of our recommender system in Section \ref{section:methodology}. The value of our system is demonstrated through experiments on different exercises in Section \ref{section:results}. Finally, we conclude by highlighting the salient points of our recommender system and identifying directions for future work, in Section \ref{sec:discussion}.

\begin{figure*}[tb]
    \centering
    \includegraphics[width=\linewidth]{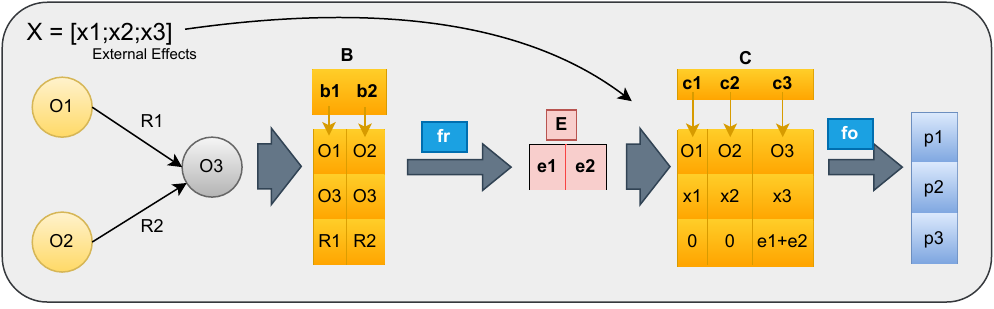}
    \Description[It shows the working of Interaction Network ]{The image shows an example of how an Interaction Network works. Given three objects, o1,o2, and o3, as nodes and two relations, R1 and R2, as edges going from o1 to o3 and o2 to o3, respectively. Using this data, function f_r evaluates effects e1 and e2 corresponding to R1 and R2. Function f_o uses these effects and object's state to predict the next state.}
    \caption{\textbf{Example showing working of Interaction Network. Vector b includes sender and receiver object details, Vector e is the resulting effect of an interaction. Vector c includes interaction effects and the exercised object.}}
    \label{fig:IN_model}
\end{figure*}

\section{Related Work}
\label{section:RelatedWork}
Several recently proposed systems \cite{ng2020posture, vyas2019pose, wang2021monocular} have used state-of-the-art pose estimation techniques to craft heuristic joint angle thresholds for pose feedback. Recently, real-time pose diagnosis was done by \citet{ alatiah2020recognizing} using pre-calculated range of motion. Similarly, \citet{ying2021aicoacher} used pre-stored correct exercises to detect incorrect moves. Such systems offer only binary feedback without any corrective recommendations.

More granular diagnoses are possible in a system recently proposed by \citet{liu2020posture}, who learned three joint angle indicators for each rep using a RNN and provided visual diagnosis for two simple upper body dumbbell exercises with high accuracy. However, their approach requires per frame annotation for training. Similarly, \citet{ gharasuie2021performance} developed a low-cost system using AlphaPose\cite{fang2017rmpe} based arm angles for upper-body exercises to count reps and also quantify exercise phase parameters to estimate user fatigue levels.
While such heuristic based methods provide helpful textual feedback in some instances, they tend to work well only for isolation arm exercises involving only few joints and do not achieve significant diagnostic accuracy without extensive frame-level annotation. Our system, in contrast, with a more sophisticated inference engine, works well for compound exercises using only video-level annotation. 

Closer technically to our approach, Pose Trainer \cite{chen2020pose} uses OpenPose \cite{cao2021openpose} on dumbbell exercises along with Dynamic Time Warping against template moves for rep diagnosis. They use angular heuristics for exercise feedback. Similarly, AI Coach \cite{wang2019ai} compares sports trajectories against pre-annotated bad poses, recommending an exemplar-based video for all identified bad pose frames. This is in contrast with our system, wherein holistic, real-time,  body-focused textual feedback is offered to users, immediately enabling them to correct any unsafe exercise posture.

\begin{table}[t]
    \caption{\textbf{Body Landmarks for four full body exercises.}}
    \label{tab:landmarks}
    \begin{tabular}{|c|c|c|c|}
        \hline
        \textbf{Squats} & \textbf{Sit-ups} & \textbf{Push-ups} & \textbf{Lunges} \\ \hline 
        Nose & Nose & Shoulder & Shoulder   \\ \hline
        Left Hip &  Shoulder & Hand & Hip \\ \hline
        Right Hip & Hip &  Hip & Front Knee   \\ \hline
        Left Knee & Knee & Toe & Back Knee \\ \hline
        Right Knee & - & - & Back Toe \\ \hline
        - & - & - & Front Heel \\ \hline
    \end{tabular}
    \Description[Landmarks for each exercise.]{The table shows the landmarks used for each exercise. For squats, it's Nose, Left hip, right hip, left knee, and right knee. For sit-ups, it's the nose, shoulder, hip, and knee. For push-ups, it's shoulder, hand, hip, and toe. For lunges, it's shoulder, hip, front knee, back knee, back toe, and front heel.}
\end{table}

\begin{figure*}[t]
    \centering    \includegraphics[width=.7\textwidth,height=.7\textwidth,keepaspectratio]{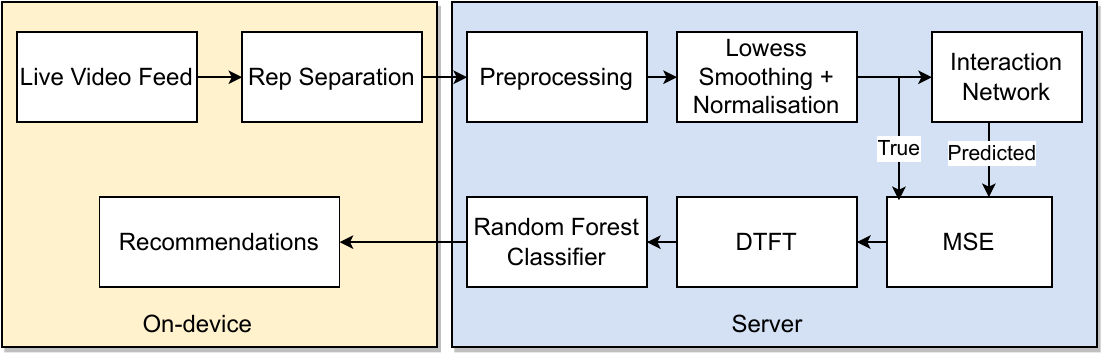}
     \Description[Flowchart of our pipeline.]{This image shows the general workflow of our pipeline. We begin with extracting body landmarks through Mediapipe API on test video. Then representative landmarks for each exercise are chosen and normalized. We feed this data to the Interaction Network to get MSE time series for the landmarks. These time series are transformed to frequency domain using DTFT and a Random Forest classifier is used to classify the test video into correct class or one of the incorrect classes with proper recommendation displayed to the user.}
     \caption{\textbf{Flowchart illustrating our Interactive System's workflow.}}
    \label{fig:pipeline}
\end{figure*}

\section{Physics prediction using Interaction Network}
\label{section:INgeneral}

Interaction Network (IN) 
\cite{battaglia2016interaction} is a Graph Networks based general-purpose physics engine simulator. The graph nodes represent the landmark joints (Table \ref{tab:landmarks}), and the edges represent the natural joint relations. The IN  (Figure \ref{fig:IN_model})  entails a relation-centric function $f_R$ for predicting interaction effects and an object-centric function $f_O$ for predicting the next step dynamics using the calculated interaction effects.
It feeds the sender landmarks’ properties($O_S$), the receiver landmark properties($O_R$), and their relational properties($R_a$) as matrices at the current time to the relation-centric function $f_R$, which outputs the effect matrix $E$, therefore we have \[E = f_R( O_S; O_R; R_a ) \:\:\:\: \textit{where ; means concatenation}.\]
Product of the effect matrix with the binary receiver matrix $R_r$ yields $\bar{E}$ = $ER_r^T$ assimilating the net effects on each receiver object in its columns. This, along with the object matrix O, is fed to the object-centric function $f_O$ to predict the next state $P$ of each landmark.
\[P = f_O(O;\bar{E}) \]
For more information, readers are requested to refer to the IN paper \cite{ battaglia2016interaction}.

\section{Proposing Pose Correcting Recommendations}
\label{section:methodology}
We first outline the overall methodology of our pipeline, followed by a detailed description of its sub-components. To begin with, we feed a recorded or live video to our pipeline, which predicts per-frame keypoints for 25 joints through Mediapipe API \cite{ bazarevsky2020blazepose}. Depending on motion evolution, we select exercise-specific landmarks (Table \ref{tab:landmarks}) followed by normalization and smoothing for physics modeling. The ML model predicts the motion rollouts for all the landmarks with visibility of only the initial rep state. Using these predictions, we calculate the Mean Squared Error (MSE) for individual landmarks and then transform them to the frequency domain for further processing, as described in subsequent sections. Essentially, we use frequency domain information from the MSE signals to classify exercise reps as either correct or incorrect (in one of the multiple predefined modes of failure) using a Random Forest multi-class classifier (Figure \ref{fig:pipeline}). Thus, our pipeline receives visual input from the user side and emits textual recommendations from the model side.

\subsection{Rep Counting using Peak Prominence}
\begin{figure}[t]
    \centering
    \includegraphics[width=.9\textwidth]{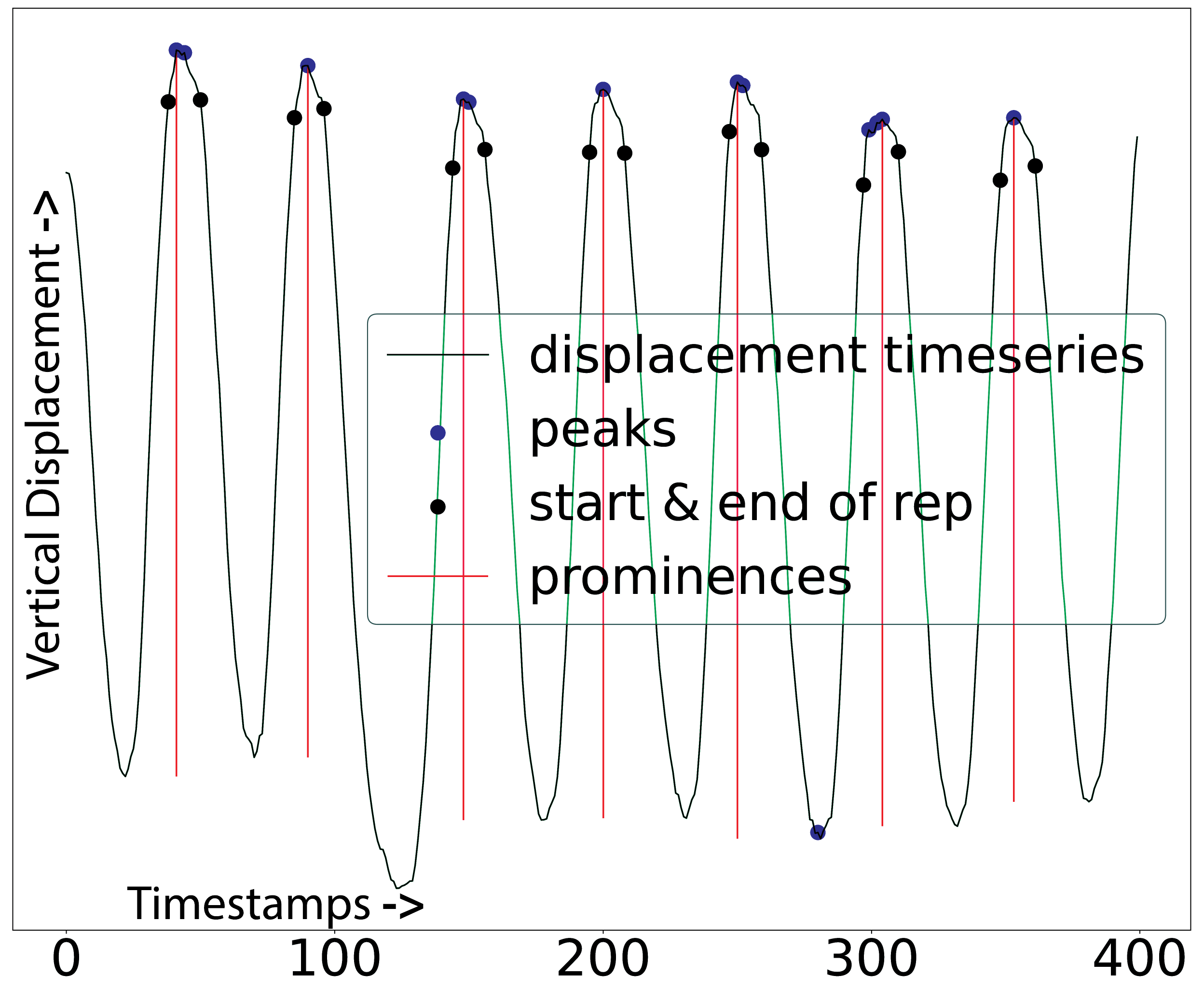}
    \Description[A plot of displacement time-series with peaks.]{This image shows a plot of vertical displacement time-series of a chosen landmark's y values, peaks, and peak prominence calculated to separate two reps and the start and end of each rep.}
    \caption{\textbf{Peak prominence over vertical periodicity for rep counting.}}
    \label{fig:rep_sep}
    \end{figure}

We exploit cyclic movements within each exercise for rep segregation. To that end, we find peaks in the periodic landmark displacement plot. 
These peaks may contain extraneous motion data, such as fragments between successive reps and other discontinuities from tired and distracted performers. To detect genuine peaks, we find peaks' importance using peak prominence and use its standard deviation as a cutoff for our high pass filter. Displacement values above the cutoff delineate the start and stop of a valid rep.Figure \ref{fig:rep_sep} shows the result of rep counting for a single lunges video.

\begin{figure}[t]

    \centering
    \includegraphics[width=.9\textwidth]{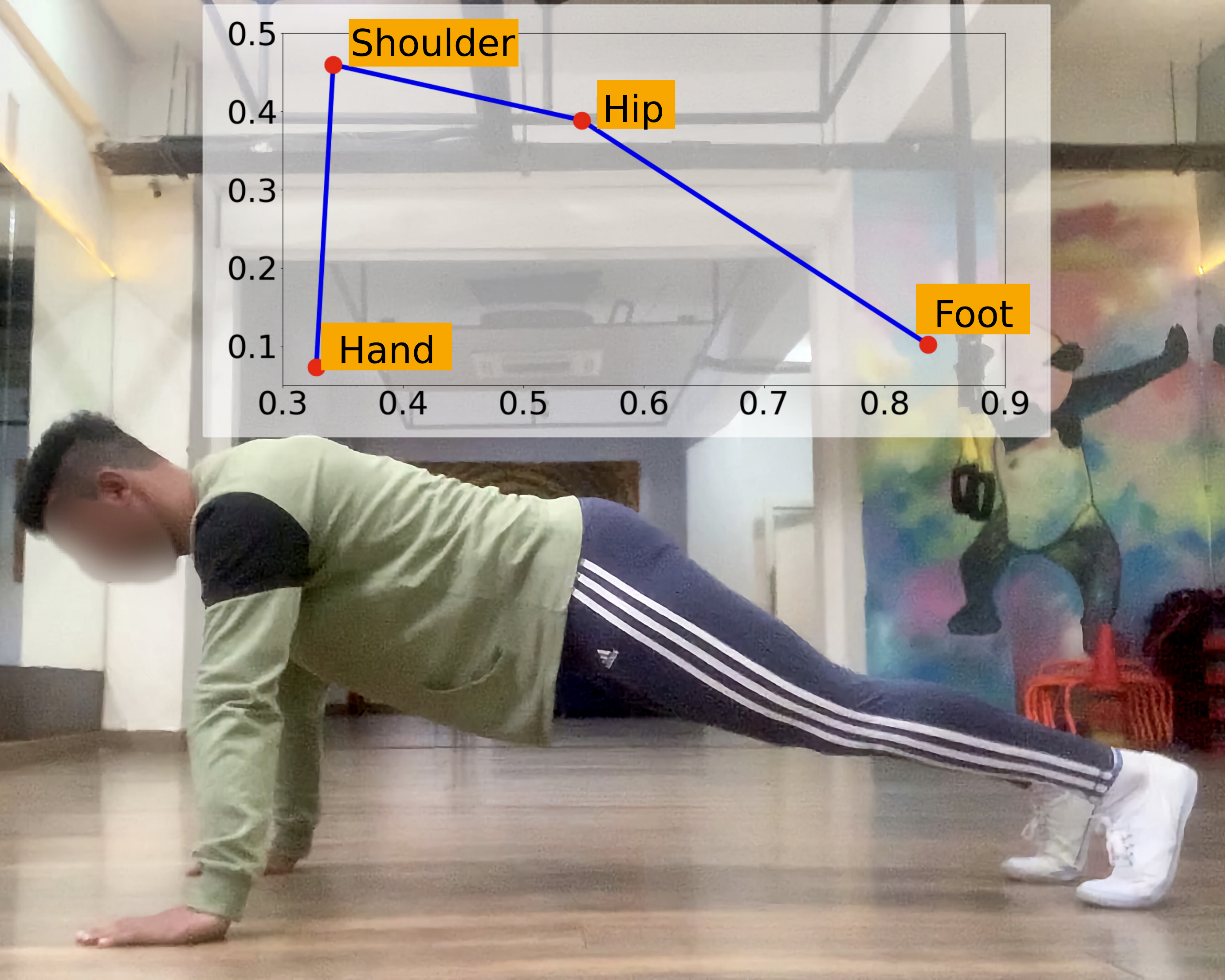}
     \Description[Figure showing an exerciser doing push-ups along with his referenced landmarks' positions]{This image shows an exerciser doing push-ups exercise. The referenced landmarks for push-ups : Hand, Shoulder, Hip, Foot are shown with a stick figure plot. }
    \caption{\textbf{Push-ups - stick figure and corresponding video frame.}}
    \label{fig:stick_fig_video}
       
\end{figure}

\subsection{Preprocessing}
MediaPipe API provides 3D positional time series data for 25 body landmarks for each exercise. We transform these coordinates for unidirectional facing and use the resulting view along with the landmark's displacement amplitude to fix the representative landmarks for an exercise. 

We apply Locally Weighted Scatterplot Smoothing (LOWESS) to each time series \cite{cleveland1981lowess} to reduce noise, discard reps with significant pose estimation errors and Min-Max normalize the coordinates to induce translational invariance, finally outputting a stick figure representations for each rep (Figure \ref{fig:stick_fig_video}).

\subsection{Learning exercise dynamics}
We represent body landmarks as graph nodes and the natural body connections as the edges, one each for the forward and backward direction. The choice of landmarks depended on our understanding of the biomechanics of each exercise. We consider only 2D position and velocity as nodes’ attributes. Relational attributes include joint-to-joint distances and angles, with positive and negative unity indicating edge direction. A top and a bottom stationary reference point is added to graph nodes with x coordinates 0.5 and y values as 0 and 1. Velocity is approximated as the difference between current and previous coordinates. 
The two feed-forward neural networks - the relation-centric model $f_R$ and the object-centric model $f_O$ consist of three and four hidden layers respectively with each layer having 256 units with ReLU activation and dropout value 0.5. The output layers have linear activation. We train the model for 2500 epochs with early stopping using AdamW optimizer {\cite{loshchilov2017decoupled}} with 1cycle learning rate policy {\cite{smith2018disciplined}}. and a learning rate of 3e-4.

\begin{table}[t]
     \centering
    \caption{\textbf{Recommendations offered for full body exercises.} }
    \begin{tabular}{|c|}
        \hline
        \textbf{Lunges}\\ \hline
        Keep your knees behind the toes  \\ \hline 
        Keep your legs closer, they are too wide apart  \\ \hline 
        \textbf{Squats}\\ \hline
        Keep your knees behind the toes  \\ \hline 
        Don't bend your knees inward  \\ \hline 
        Keep your Feet shoulder-width apart   \\ \hline
        \textbf{Situps}\\ \hline

        Your back should rise up completely \\
        \hline
        \textbf{Pushups}\\ \hline

        Keep your Knees-hips-Shoulders in a straight line  \\ \hline
        Lower your chest to align it with hip  \\ \hline
         Lower your hips  \\ \hline
         Your chest should not touch the ground  \\ \hline
        
    \end{tabular}
    \Description[Sample of recommendations our pipeline provides]{The table shows the recommendations we provide for different anomalies for Lunges and Squats label.}
    \label{tab:dataset_stats}
\end{table}

\begin{table}[t]
    
    \caption{\textbf{F1 scores of training with different classifiers.}}
    \centering
    \begin{tabular}{|c|c|c|c|c|}
        \hline
        Classifier & Squats & Push-ups & Lunges & Sit-ups\\ \hline

        SVM & 0.81 & 0.96  & 0.93  & 0.97  \\ \hline
       KNN & 0.75 & 0.87  & 0.94 & 0.92  \\ \hline
         Naive Bayes & 0.74  & 0.95 & 0.92  & 0.98 \\ \hline
        Logistic Reg. & 0.85 & 0.97  & 0.95& 0.98  \\ \hline
         LDA  & 0.87  & 0.98  & 0.97  & 0.97  \\ \hline
      Random Forest & \textbf{0.94} & \textbf{0.98} & \textbf{0.97} & \textbf{0.98} \\ \hline     
    \end{tabular}
   
    \Description[Performance with different classifiers]{This table shows the performance of our pipeline with different classifiers using weighted F1 scores and standard deviations over five training runs. We tested six classifiers - Support Vector Machine, K Nearest Neighbours, Naive Bayes, Logistic Regression, Linear Discriminant Analysis, and Random Forest. Random Forest gave the best results in all cases, occasionally matched by one of the other classifiers. }
    \label{tab:classifiers_table}

\end{table}

\begin{figure*}[tp]
    \centering
    \includegraphics[width=\linewidth]{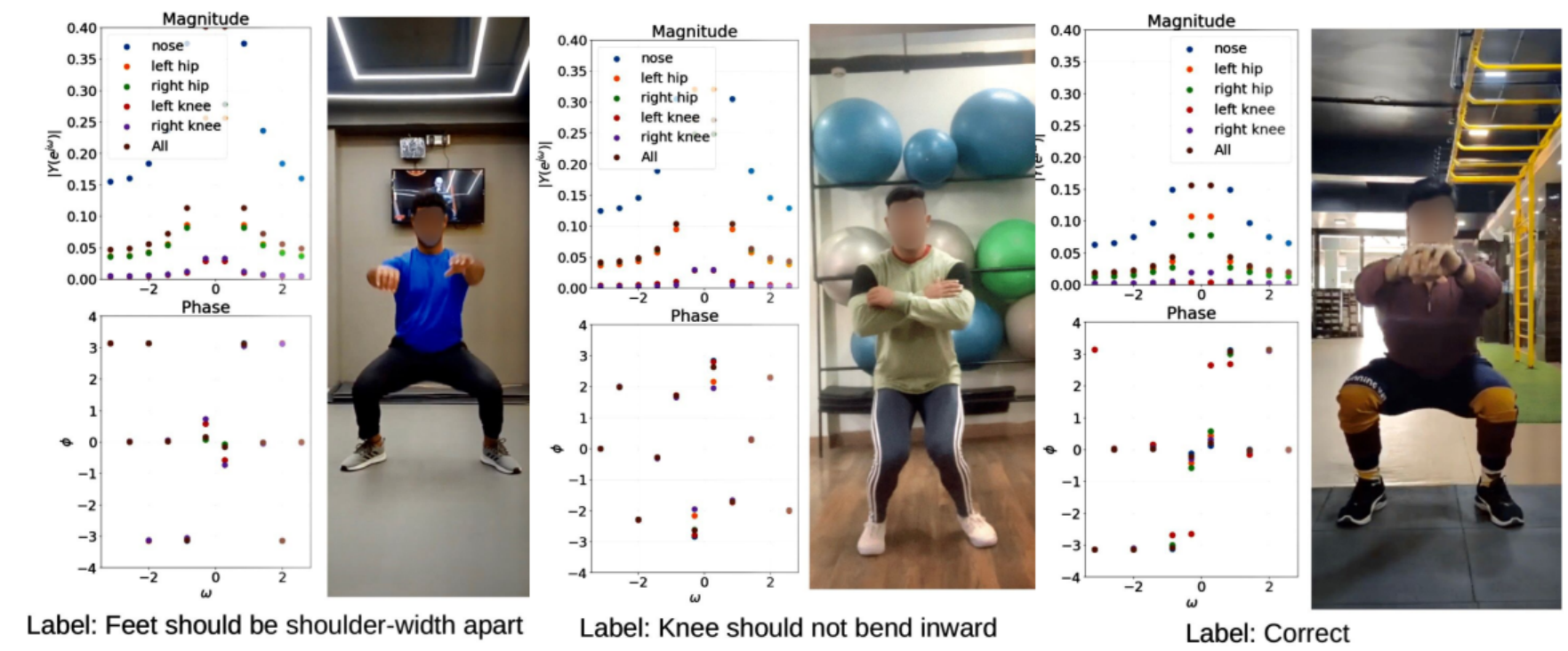}
    \caption{\textbf{DTFT representations of error signature of different squats labels. Note how the phase plots for the incorrect squats differ from each other as well as from a correct squat systematically.}}
    \label{fig:prominence}
    \Description[DTFT error representation for the different incorrect labels of squats.]{The figure shows three images, and each image shows two parts - the DTFT error signature of squats on the left and the exerciser on the right. Three images have squats with incorrect labels of "feet should be shoulder width apart" and "knee should not bend inward" and there is one correct exercise with the label "Correct".}
\end{figure*}

\subsection{Error Analysis and Rep Classification}

The MSE time series emitted by the IN informs the subsequent stages. We hypothesize that our physics engine predicts the correct method of exercising, such that considerable deviation from it would hint at an incorrect rep. Further, the specific combination of MSE from different body components would hint at the precise mistake made by the exerciser.

To extract this information, we transform the MSE time series from all the representative landmarks to the frequency domain using the discrete-time Fourier transform (DTFT). DTFT provides magnitude and phase values for each time series. Conversion to the frequency domain helps in two ways. It gives a fixed-sized representation of the variable-length time series. It also helps to extract the features of the time series. This output of the DTFT, called the error signature (Figure \ref{fig:prominence}), is a vector representation of an exercise rep of variable duration. For our case, we take the principal 11 amplitudes and the corresponding phase values to build the error signature of each exercise rep. At the final stage of our pipeline, we use a Random Forest classifier for classification, operated in a multi-class classification setting, with recommendation labels used as shown in Table \ref{tab:dataset_stats}. We found Random Forest was the most consistent classifier across all the exercises. (Table \ref{tab:classifiers_table}) and tuned its hyperparameters using randomized search cross-validation

\begin{table}[t]
  \caption{\textbf{Baselines comparisons for four full body exercises (left) and two upper body exercises(right). Classification results reported using weighted F1 score with standard deviations over five train-test runs (*Shoulder Press results reported for two incorrect classes).}}

  \begin{minipage}[t]{.58\textwidth}
    \begin{tabular}{|c|c|c|c|c|}
        \hline
        Model & Squats & Push-ups & Lunges & Sit-ups \\ \hline
        MLP & 0.91 $\pm$ 0.02 & 0.98 $\pm$ 0.03 & 0.95 $\pm$ 0.03 & \textbf{0.99 $\pm$ 0.01} \\ \hline
        RNN & 0.85 $\pm$ 0.04 & 0.98 $\pm$ 0.01 & 0.94 $\pm$ 0.01 & 0.98 $\pm$ 0.02 \\ \hline
      GRU & 0.87 $\pm$ 0.03 & 0.98 $\pm$ 0.01 & 0.93 $\pm$ 0.02 & 0.94 $\pm$ 0.04 
      \\ \hline
         
         IN & \textbf{0.94 $\pm$ 0.02} & \textbf{0.98 $\pm$ 0.01} & \textbf{0.97 $\pm$ 0.01} & 0.98 $\pm$ 0.01 
 \\ \hline
     
    \end{tabular}

    \label{tab:baselines_full}
  \end{minipage}%
  \begin{minipage}[t]{.42\textwidth}

    \centering
    \begin{tabular}{|c|c|c|}
        \hline
        Model & ShoulderPress* & FrontRaise\\ \hline
       Ng \cite{ng2020posture} & 0.90 & 0.77 \\ \hline
       
       PoseTrainer\cite{chen2020pose} & 0.49 & 0.76  \\ \hline

        MLP & \textbf{0.99 $\pm$ 0.01 }&  0.82 $\pm$ 0.04 \\ \hline
        RNN & 0.99 $\pm$ .01 & 0.79 $\pm$ .05\\ \hline
      GRU & 0.95 $\pm$ .06 & 0.80 $\pm$ .04 
      \\ \hline
         
         IN & 0.98 $\pm$ 0.01 &\textbf{ 0.88 $\pm$ 0.03}
 \\ \hline
    \end{tabular}

    \label{tab:baselines_upper}
  \end{minipage}

     \label{tab:baselines}
      \Description[Weighted F1 scores for baselines and IN]{The Table shows F1 scores for GRU, NN, RNN and IN for 6 exercises - Squats, Push-ups, Lunges, Sit-ups, Shoulder Press, Front Raise. IN performs best for Squats,Push-ups, Front Raise and Lunges and comparatively for Sit-ups and Shoulder Press.}
\end{table}
\section{Empirical Evaluation}
\label{section:results}

\begin{figure*}[t]
    \centering
    \includegraphics[width=\textwidth]{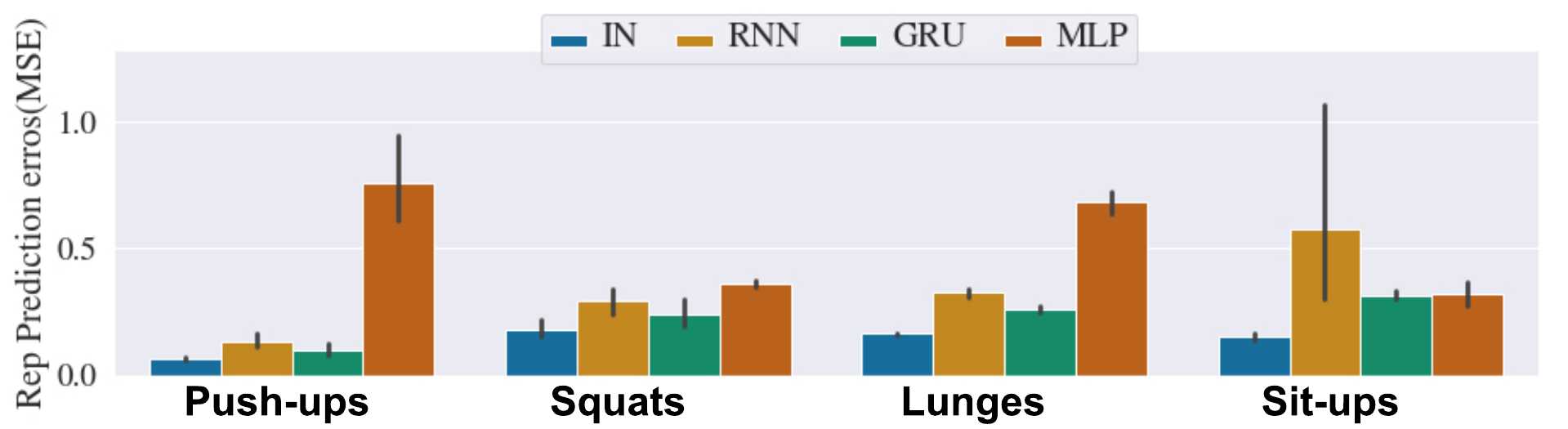}
    \caption{\textbf{Average rollout prediction errors over exercise reps(MSE) for Baseline Models and Interaction Network. Even though MLP have good F1 scores (Table \ref{tab:baselines_full}) in some cases, high prediction error makes their performance unreliable.}}
    \label{fig:pred_error_baseline}
    \Description[MSE for baselines and IN]{The image shows MSE scores for GRU, NN, RNN and IN for four exercises - Push-ups, Squats, Lunges, and Sit-ups. IN has least MSE for all cases, showing that it can accurately learn the physics of the exercise.}
\end{figure*}
\subsection{Data}

For the evaluation of full-body exercises, we used a proprietary dataset from E-Trainer Analytics Wizard Pvt. Ltd. This dataset contains the front and side view of seven exercisers performing more than 150 reps for four exercises - squats, push-ups, lunges, and sit-ups. Each exercise has one correct class, whereas incorrectly done exercises could belong to multiple classes.
Incorrect videos were annotated with corrective recommendations by expert physical trainers.  

We used a train test split of 60\%-40\% to train the classifier on incorrect classes. Similarly, 60\% of the correct class data was used to train our physics engine.
To compare against existing approaches of form prediction, we conducted evaluation using annotated data for shoulder press and front raise from a publicly available dataset~\cite{ng2020posture}

\subsection{Baseline Comparisons}
Among available learnable dynamics predictors,  we exploit IN for their interpretability and simplicity. Our pipeline can also function with other motion predictors to the degree that they can accurately mimic the dynamics of the exercise. To test this hypothesis, we evaluate our model against several baselines.

The \textbf{Multi layer perceptron (MLP) Baseline}, with three 256-length hidden layers and ReLU activation, has all information to learn the interaction dynamics, requiring it to assimilate relation indices implicitly without explicit scene factorization.
The \textbf{Recurrent Neural Network (RNN) and Gated Recurrent Unit (GRU)}, capable of modeling posture evolution, have three recurrent units with three features in the hidden state. The final hidden state output is fed to a fully connected layer to predict future dynamics. All the above baselines use flattened node attributes as input.

Additionally, we compare our pipeline against popular heuristic techniques (\cite{ng2020posture}, \cite{chen2020pose}), which examine reps using geometric thresholds over joint features. Next, we describe ablation modifications on IN's architecture and input.

The \textbf{Attribute Hidden IN} uses the same IN architecture but with an empty relation attribute matrix which, in principle, could be deduced from position data demanding estimation of complex distance and inclination functions. The \textbf{Independent object IN} simulates removing the relation-centric component by zeroing out the interaction effects, which incapacitates it from modeling object-object interactions.
The \textbf{Fully Connected (FC) IN} connects each joint with every other joint, procuring the same capacity as the IN but involving additional irrelevant inputs. The \textbf{Global Connection (GC) IN}, additionally, connects all the landmark points to the two stationary reference points, modeling both local and global interactions for superior information propagation.

\subsection{Results}

We evaluate our diagnostic system on six exercises using three criteria: Rep Counting , Posture diagnosis, and Real-time prediction. Our peak-prominence based algorithm perfectly counted all the reps in the full-body exercises. For each rep detected, we measured recommendation accuracy using weighted F1 scores in a multi-class classification setting.

\subsubsection{Posture diagnosis}
Our results (Table  \ref{tab:baselines}) indicate that a pipeline endowed with physics learning capability outperforms all baselines effectively differentiating correct and incorrect exercise reps. 
Performance can be comparable for fewer prediction classes (e.g., Sit-ups, Shoulder Press) or simple exercises with a small range of joint motion(e.g., Push-ups). However, the performance gain is evident as the number of incorrect classes increases (Front Raise - Table \ref{tab:variable_data_size}).

Since classification results are an indirect measure of physics modeling, we also explore the MSE for the next state prediction (Figure  \ref{fig:pred_error_baseline}). In all cases, a physics learning engine best describes motion dynamics. Even though the MLP showed good classification (Table \ref{tab:baselines}), it significantly deviates from the actual exercise, causing deviant performance, especially with increasing exercise complexity (Table \ref{tab:variable_data_size}).  Even an ill-suited model can classify well if the error signatures are separable. However, such arbitrary performance gains do not scale well as the number of classes increases.

\begin{figure*}[t]
    \centering
    \includegraphics[width=\textwidth]{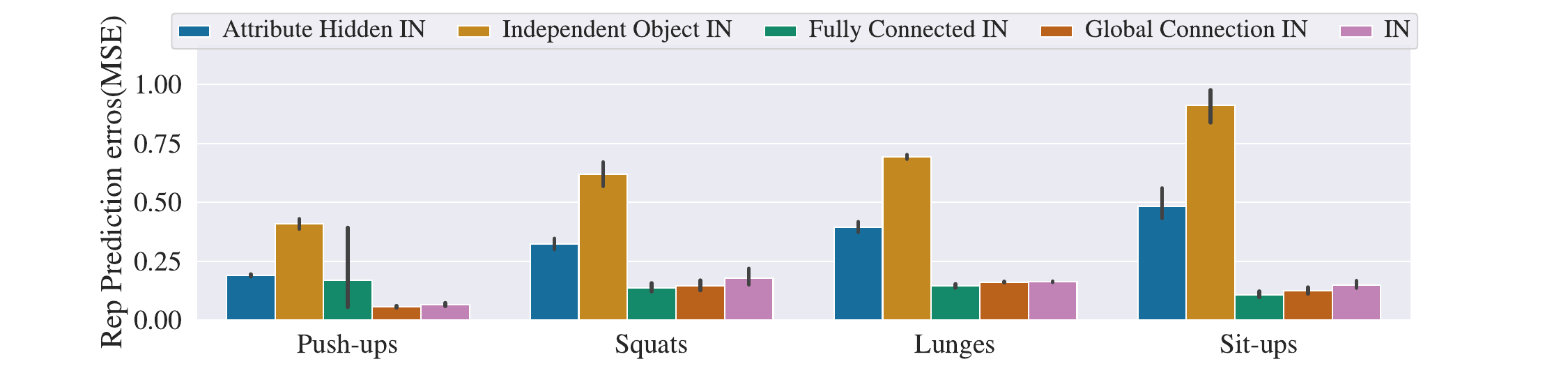}
    \caption{\textbf{Average rollout prediction error over exercise reps(MSE) for Ablation Models and Interaction Network. Models without relation information experience significant drop in performance for dynamics prediction.}}
    \label{fig:pred_error_abl}
    \Description[MSE values for ablation models and IN]{The image shows MSE for 4 ablation model - Attribute Hidden IN, Independent Object IN, Global Connection IN and Fully Connected IN. Attribute hidden IN and Independent Object IN perform poorly as they do not have Interaction attributes and relation information respectively. Other ablation models have similar performance.  }
\end{figure*}

Our ablation study on IN's performance gain indicates that relational attributes are critical for learning the physics of interactions (Figure \ref{fig:pred_error_abl}). Reasonable variations in the joint-to-joint links (as with FC-IN and GC-IN) show similar effects, with GC-IN edging slightly over the vanilla IN, probably due to faster information propagation. Stochastic interactions with involuntary factors (like fatigue and distractions) possibly contain the expected performance gain from global propagation. The FC-IN with redundant information in its irrelevant relations competes well, presumably because the IN learns to weigh the importance of exercise-specific relations. This architecture may, thus, bypass the requirement of an explicit relational matrix, provided that the pose estimation accurately detects all body landmarks. As expected, the independent object IN without relation function $f_r$ finds it challenging to model the motion dynamics. Similarly, the Attribute Hidden IN is unable to exploit the joints' information without relational attributes.

\begin{table}

    \caption{\textbf{Baselines Comparison for Front Raise with increasing number of classes . Methods with poor dynamics modeling show significant performance drop.}}
    \centering
    \begin{tabular}{|c|c|c|c|}
        \hline
         FrontRaise & 2 Classes & 4 Classes & 6 Classes   \\ \hline
         MLP & 0.96  & 0.91  & 0.82  \\ \hline
         RNN & 0.93   & 0.90 & 0.79  \\ \hline
         GRU & 0.93  & 0.89  &  0.80   \\ \hline
         IN & 0.96  & 0.91   &  0.88  \\ \hline    
    \end{tabular}
    \Description[Baseline comparison with Increasing Exercise Complexity in Front Raise]{This table shows that as the complexity of exercise increases, the performance decreases. The performance drop is least for IN when we increase number of classes from 2 to 6 and most for MLP which had the highest MSE among all baselines.}
    \label{tab:variable_data_size}

\end{table}

\subsubsection{Diagnosis latency }
The MediaPipe Android API, fed with the exercise's camera feed, outputs the joints' coordinates time series. After rep segregation, each rep data is passed to the server, where our pipeline classifies it as correct or diagnoses it as a mistake of a particular type. A corrective recommendation specific to the estimated diagnosis is displayed to the user through our mobile application. For exercisers operating at normal tempo, this feedback arrives before their next rep is halfway complete prompting the user to instantly correct any mistakes in technique (see Table \ref{tab:rep_del} for a quantitative summary and Table \ref{tab:links} for examples of live recommendations videos).

\begin{table*}[tb]
  \begin{minipage}[t]{.45\textwidth}

 \caption{\textbf{Lag time(seconds) for new rep recognition.}}

    \centering
    \begin{tabular}{|c|c|c|}
        \hline
        Exercise & Mean(sec) & Standard deviation(sec)
        \\ \hline
        Squats & 0.55 & 0.13 \\ \hline
        Sit-ups & 0.39 & 0.07 \\ \hline
        Push-ups & 0.36 & 0.11 \\ \hline
        Lunges & 0.54 & 0.09 \\ \hline
    \end{tabular}
    \label{tab:rep_del}
    \Description[Lag time in new rep recognition]{This table shows the time lag between the user finishing a rep and its detection. As the rep is detected it is sent to our server for labelling and appropriate suggestions is conveyed to the user.}
    \end{minipage}
  \begin{minipage}[t]{.45\textwidth}

    \caption{\textbf{Link to videos of interaction network predictions rolled out over time and of exercise sessions diagnosed using our system in real-time.}}
    \centering
     \begin{tabular}{|c|c|}
         \hline
          Data type & Link \\ \hline
         IN predictions simulations & \href{https://drive.google.com/drive/folders/1VCilOlQZUrbffUhTQUOfA1ZyvSG8HvDB?usp=sharing}{\color{blue}{Link}} 
         \\ \hline
         Exercise Demo & \href{https://drive.google.com/drive/folders/1CVgKwDqE8jgowA0JFKp4SY1uncaP-v1h?usp=sharing}{\color{blue}{Link}}
         \\ \hline     
    \end{tabular}
     
     \label{tab:links}
     \Description[Links to different video folders.]{The table shows the links to interaction networks prediction simulations and the exercise demo.}
 \end{minipage}

\end{table*}

\section{Discussion and Limitations}
\label{sec:discussion}

Self-training is gaining popularity as more and more people feel disinclined to commit to a dedicated gym routine. By providing accurate real-time recommendations, we can benefit such users without compromising their day-to-day schedule. Our recommender system focuses on rep counting and diagnosis, assuming that the exercise performed is known (or is easily knowable). For instance, \citet{moran2022muscle} used MediaPipe API \cite{ bazarevsky2020blazepose} for pose recognition to detect the type of exercise performed in real-time, a capability that could easily inform exercise type in our pipeline. 

Thus, to summarize, this paper offers a novel system for recommending form corrections to exercisers performing rep-based training in real-time with high precision. We introduce the use of learnable physics engines to model body physics, a task for which they are very well-suited. The success of our physics model permits downstream classifiers to accurately diagnose modes of failure of exercises using differential prediction error residuals between the model prediction and actual observations. Empirical evaluations show that our system diagnoses defective techniques in complex full-body exercises with high sensitivity and specificity. We expect the adoption of such interactive systems to help healthcare providers scale up access to supervised physical exercise.

We conclude with a brief exploration of the limitations of our system, and possible directions for future work. The most critical technical limitation of the present system is its reliance on pre-defined relational attributes for each exercise's Interaction Network. These attributes depend on the nature of human biomechanics and must be decided beforehand. Learning relational attributes from data could improve this performance even further, a clear direction for future work. Our system is currently tested only for exercises with significant vertical periodicity, an artifact of our peak-prominence based rep-counting scheme, though vertical periodicity also exists in many other exercises. Replacing this with a more sophisticated rep-counting method could extend our system's capabilities to a more general set of exercises. In particular, given the known diagnostic value of gait analysis in predicting health outcomes for the elderly~\cite{cesari2005prognostic, verghese2009quantitative}, extending this system's digital diagnostic capabilities to monitoring and diagnosing gait-related problems presents a very promising direction for future work. 
\bibliographystyle{ACM-Reference-Format}
\bibliography{ref}
\end{document}